\documentclass[journal]{IEEEtran}

\usepackage[vlined, ruled, linesnumbered, commentsnumbered]{algorithm2e}
\usepackage{amsmath}
\usepackage{cite}
\usepackage{color}
\usepackage{xcolor}
\definecolor{chred}{rgb}{0.8,0,0}
\definecolor{chgray}{rgb}{0.5,0.5,0.5}
\usepackage{graphicx}
\usepackage{caption}
\usepackage{dblfloatfix}
\usepackage{subfigure}
\usepackage{eqlist}
\usepackage{txfonts}
\usepackage{url}
\usepackage{footmisc}
\usepackage{booktabs}

\usepackage{multirow}
\usepackage{threeparttable}
\newcommand{\bb}[1]{\textcolor{blue}{#1}}
\newcommand{\br}[1]{\textbf{\textcolor{magenta}{#1}}}
\newcommand{\brt}[1]{\textcolor{magenta}{#1}}
\newcommand{\bg}[1]{\textcolor{brown}{#1}}
\newcommand{\gr}[1]{\textcolor{gray}{#1}}
\newcommand{\bsf}[1]{\textbf{\textsf{#1}}}

\begin{document}
\bstctlcite{IEEEexample:BSTcontrol}
\title{Regrasp Planning using 10,000s of Grasps}

\author{Weiwei~Wan,~\IEEEmembership{Member,~IEEE,} and
        Kensuke~Harada,~\IEEEmembership{Member,~IEEE}
\thanks{Weiwei Wan and Kensuke Harada are with National
Institute of Advanced Industrial Science and Technology (AIST), Japan. Kensuke
Harada is also affiliated with Osaka University, Japan.
{\tt\small wan-weiwei@aist.go.jp}}}

\markboth{Journal of \LaTeX\ Class Files,~Vol.~x, No.~x, xxxx~2016}%
{Shell \MakeLowercase{\textit{et al.}}: Bare Demo of IEEEtran.cls for IEEE Journals}

\maketitle

\begin{abstract}
This paper develops intelligent algorithms for robots to reorient objects. Given
the initial and goal poses of an object, the proposed algorithms plan a sequence
of robot poses and grasp configurations that reorient the object from its
initial pose to the goal. While the topic has been studied extensively in
previous work, this paper makes important improvements in grasp planning by
using over-segmented meshes, in data storage by using relational database, and
in regrasp planning by mixing real-world roadmaps. The improvements enable
robots to do robust regrasp planning using 10,000s of grasps and
their relationships in interactive time. The proposed algorithms are validated
using various objects and robots.
\end{abstract}

\begin{IEEEkeywords}
Grasp Planning, Manipulation Planning, Reorient Objects, Preparatory Planning
\end{IEEEkeywords}

\IEEEpeerreviewmaketitle

\section{Introduction}

\begin{figure*}[!htbp]
	\centering
	\includegraphics[width=\textwidth]{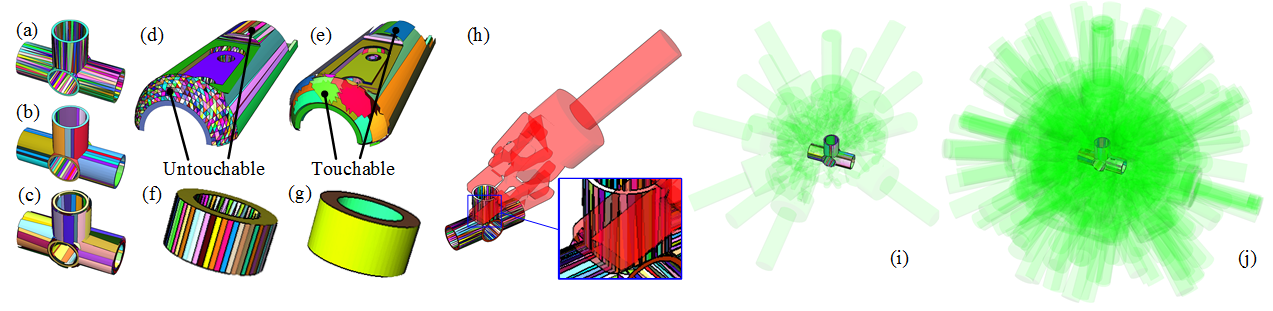}
	\caption{(a)-(b) Results of conventional segmentation algorithm using two
	different threshold (see $\tau$ in Fig.\ref{oversegalg}).
	(c) Result of the over-segmentation algorithm. (d) Conventional
	segmentation algorithm leads to uneven facets and untouchable small facets.
	(e) Compared with (d), over-segmentation algorithm overcomes the disadvantages.
	(f)-(g) Bad $\tau$ (Fig.\ref{oversegalg}) leads to many small facets ((f)) or one large
	curved facet ((g)). (h) Grasp planning fails at small facets since they are
	not large enough to support finger pads.
	(i)-(j) The grasp planner using over-segmentation finds more grasps ((j)) than
	conventional segmentation ((i)).}
	\label{overseg}
\end{figure*}

\IEEEPARstart{T}{his} paper develops intelligent algorithms for robots
to reorient objects. Given the mesh models of objects, their initial and goal
poses, and the kinematic and dimensional parameters of robots, the algorithms
developed in the paper could find a sequence of robot poses and grasp
configurations that reorients the objects from their initial poses to the goals.

The algorithms developed include grasp planning algorithms, placement planning
algorithms, and graph searching and motion planning algorithms.
Developing these intelligent algorithms are important to industrial robots. In
factories, objects are sent to robots in boxes. The objects are in various
poses. A robot is required to recognize the objects, pick them
up, and reorient them to specific poses for particular use. Examples include:
(1) Packing items. Each item should be reoriented to the same orientation and
packed into a box.
(2) Assembly. Each part in the assembly should be reoriented to specific poses
to fit others. (3) Using tools. A tool should be reoriented to have its
tool center point facing targets. These tasks require a robot to be
equipped with both high-precision vision systems and robust grasp and
manipulation planning systems. This paper studies the grasp and manipulation
planning systems. It develops robust algorithms to do regrasp by using
10,000s of auto-planned grasps.

The developed algorithms include: (1) A grasp planner which automatically plans
available grasp configurations using the mesh models of objects. (2) A placement
planner which automatically plans stable poses of an object on a planar
surface (table top). (3) A regrasp planner which builds and searches regrasp
graphs to generate a sequence of robot poses and grasp configurations that
reorients objects from their initial poses to the goals.

While these algorithms have been studied for thirty years and also
have been extensively discussed and re-developed in several of our previous
work, this paper makes important improvements to make them robust. It leverages
the computational ability of modern computers to deal with various robots,
objects, and combinatorics.
The main contributions are as follows. (1) We propose a grasp planning system
which uses over-segmented mesh surfaces to sample contact points. The over-segmented
surfaces provide more even segments and robuster measurement of
contact regions. (2) We employ RDB (Relational DataBase) to manage the large
amount of data generated by planning algorithms. RDB makes it easy to maintain the
relationships among grasp configurations, placements, objects, and robots. It
enables saving and retrieving gigabyte-level data to build regrasp graphs
and select grasps and placements all over a table in front of a robot. (3) We
build the regrasp graph like a roadmap in a robot's workspace and search the graph to find a sequence of robot
poses and grasp configurations that reorients the objects from their initial
poses to the goals. The graph, together with contributions (1) and (2), makes it
possible for different robots to reuse 10,000s of grasps and their relationships
to reorient objects with various initial and goal poses in interactive time.

\section{Reorienting Objects using Regrasp Planning}%


The seminal work that studied reorienting objects using regrasp is
\cite{Pierre87}.
The work motivated many researchers.
\cite{Rohrdanz97}\cite{Hajime98}\cite{Sascha99}\cite{Cho03} are some of
the early publications that applied similar technique to various robots and
grippers. These early work concentrated on the regrasp aspect. Their grasp
and motion planning was limited by the computational capacities at that
time. The number of grasps were small and the grasp planning was based on block
models, primitive matching, or manually selected values.

More recent work involved better grasp and motion planning. For example,
Xue et al. \cite{Zhixing08} used shape primitives \cite{Miller03} to plan
grasps for a cup and implemented the regrasp and reorient planning of the cup using multi-finger
hands. Saut et al. \cite{Jean10} used decomposition to plan grasps and
implemented dual-arm regrasp of complicated models. King et al. \cite{King2013}
used regrasp planning to do preparatory reorient. They implemented
primitive-based prehensile and non-prehensile grasp planning
to prepare for optimal motion planning.
Simeon et al. \cite{Thierry04}
presented a framework which integrated motion planning and transfer-transit
regrasp. Hauser et al. \cite{Kris10} made a concrete description of
multi-modal motion planning and presented several implementations. Yoshida et
al. \cite{Eiichi10} applied regrasp and motion planning to a humanoid robot that
transported a box. 
Cohen et al.
\cite{Benjamin10} developed algorithms to sequentially handle an object using
several manipulators and regrasp. Nguyen et al. \cite{Nguyen16} developed
algorithms for a WALK-MAN robot to reorient an electric drill, using some
carefully selected grasps to make the manipulation robust. Chang et al. \cite{Chang08} studied the
preparatory grasps of human beings and used non-prehensile re-grasp (pushing) to
reorient pans and pots.
Lertkultanon et al. \cite{Lert16} presented an integrated regrasp and motion
planning system to reorient furniture parts.
Their grasps were based on box primitives. Krontiris et al. \cite{Krontiris15}
developed algorithms to rearrange objects. Their focus was on the high level
planning of manipulation sequence.
Similarly, Jentzsch et al. \cite{Jentzsch15} used regrasp to solve multi-modal
pick-and-place problems. Lee et al. \cite{Lee15} also used non-prehensile grasp
to plan sequential manipulation and reorient.

The essential algorithms in reorienting objects using regrasp include: (1) A
grasp planner, (2) a placement planner, and (3) a regrasp planner.
The grasp planner plans a redundant number of available grasps. The placement
planner finds stable placements of the object in the environment. It also
associates the grasps found by the grasp planner to the stable placements.
Following the grasp planner and placement planner, the regrasp planner builds a
regrasp graph by considering the shared grasps of the stable placements,
connects the initial and goal poses to the graph by solving inverse kinematics
and detecting collisions, and plans a sequence of robot poses and grasp
configurations to reorient the object from its initial pose to the goal.

This paper makes improvements on grasp planning and data management. It proposes
an improved grasp planning system using over-segmented facets, employs
relational database to manage the large amount of data generated by planning
algorithms, and builds and searches regrasp graphs like a roadmap in a robot's
workspace using the saved data and
their relationships. These improvements make it possible for a robot to reuse
10,000s of grasps and relationships to reorient objects with various
initial and goal poses in interactive time. To our best knowledge, it is the
first work that reorients objects using such a large amount of data.

\section{Grasp Planning using Over-segmented Surfaces}

\begin{figure}[!htbp]
	\centering
	\includegraphics[width=.487\textwidth]{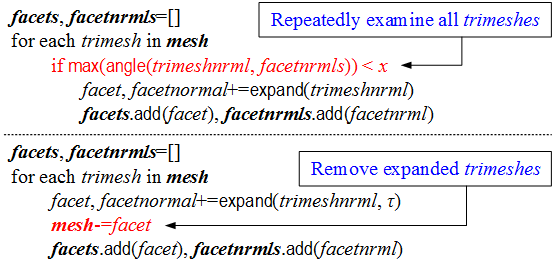}
	\caption{Upper part: The over-segmentation algorithm.
	Lower part: The conventional segmentation algorithm. The over-segmentation
	algorithm repeatedly examines all triangles in the mesh. One triangle could
	be clustered to several facets. In contrast, the conventional algorithm
	clusters a triangle to a single facet, leading to uneven segments.}
	\label{oversegalg}
\end{figure}

\textbf{Over-segmentation:} We plan grasps by over-segmenting
an object mesh into redundant number of facets. The pseudo code of the
over-segmentation algorithm is shown in the upper part of Fig.\ref{oversegalg}.
Compared with the conventional algorithm shown in the lower part, the
over-segmentation algorithm allows overlap between facets by repeatedly
examining all triangle meshes. The conventional algorithm removes the expanded triangle
meshes during segmentation, leading to uneven segmentation.

Fig.\ref{overseg} compares some results. Fig.\ref{overseg}(a) and (b)
are segmented using a conventional segmentation method which does not allow
overlapped facets. (a) and (b) are the results of two different thresholds
(parameter $\tau$ in Fig.\ref{oversegalg}).
Fig.\ref{overseg}(c) is the result of the proposed over-segmentation method. The
conventional segmentation method has several disadvantages: (1) It is
difficult to judge if a facet is safe to touch or not.
Take Fig.\ref{overseg}(d) and (e) for example. In Fig.\ref{overseg}(d), a curved
mesh surface is segmented into several flat facets using the conventional segmentation
method. Some of them are large, some others are small. Once triangle meshes are
segmented to adjacent facets, the remaining ones will be small. Large facets are
touchable, but small ones might be either touchable or untouchable. They are
segmented independently from surrounding facets and their real touchability is
difficult to judge. (2) It is difficult to tune the parameter $\tau$ which is
used to decide coplanar triangle meshes. A large $\tau$ may result into
non-planar facets. For example, the whole cylindrical surface in
Fig.\ref{overseg}(f) is mistaken as one facet. It is not planar. In contrast, a
small $\tau$ results into many small facets in Fig.\ref{overseg}(g), making it
difficult to do surface sampling and compute parallel facet pairs. (3) The
conventional segmentation method deteriorates the performance of grasp
planning. If a facet is too small to support a finger pad, there will be no
available grasps to grip at that facet, leading to fewer automatically planned
grasps and low success rate during regrasp. For example, the facets in
Fig.\ref{overseg}(h) are segmented by the conventional segmentation method. They
are too small to support a finger pad and no grips on the facet is available (the
red hand in Fig.\ref{overseg}(h) indicates an unavailable grasp). For this
reason, the number of planned grasps using the conventional method is much smaller compared to the
over-segmentation method (Fig.\ref{overseg}(i) vs. Fig.\ref{overseg}(j)).
Considering these disadvantages, we propose the grasp planning using
over-segmentation. Since the facets are over-segmented, they include the
information of a local region. Planning grasps using the over-segmented facets
are more robust and complete.

\begin{figure}[!htbp]
	\centering
	\includegraphics[width=.47\textwidth]{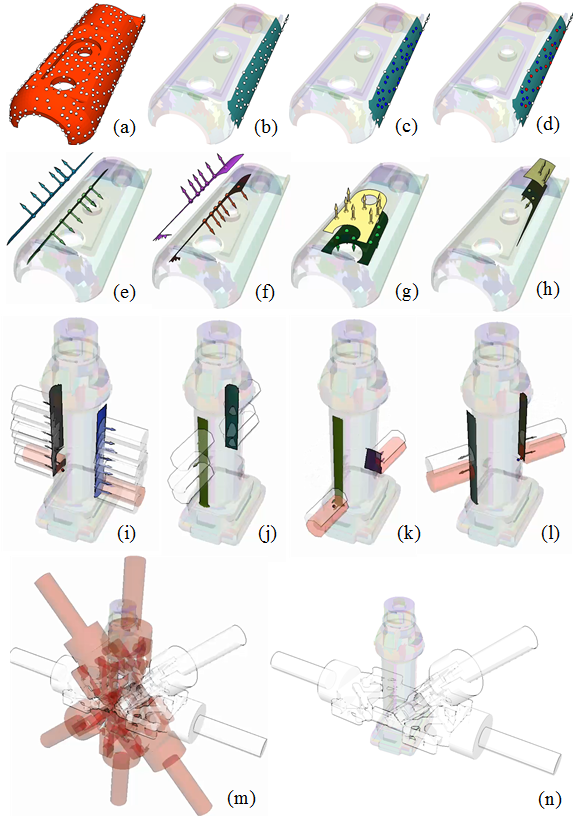}
	\caption{(a) Surface sampling. (b) Distributing samples to a facet. (c)
	Distance filter applied. White points are removed. (d) Near-neighbour filter
	applied. Blue points are merged to red points.
	(e)-(h) Some results of the parallel facets. (i)-(l) Some results of the first level
	collision detection. (m)-(n) Some results of the second level collision
	detection. In (i)-(n), the collided configurations are shown in red.}
	\label{graspplanning}
\end{figure}

\textbf{Mesh sampling:}
The next step is to sample some contact points on the facets. While
probabilistically sampling points on each over-segmented facet is an intuitive
method, it leads to redundancy. To avoid redundancy and repeatedly
computing force-closures and performing collision detections, we pre-sample on
the whole mesh surface and distribute the pre-sampled results to over-segmented
facets. One sampled point could be distributed to multiple facets. In this way,
we use one sample process to generate contact points for all over-segmented facets.
Fig.\ref{graspplanning}(a) and (b) show the results of the surface sampling and
the sampled points distributed to one over-segmented facet.

The sampled points are further refined using the following filters. (1) Distance
filter. A sampled point must neither be too near to nor too far from the
boundary of the facet. If a sampled point was too near to the
boundary, the finger pad touching that point might be on the edges of the
object, leading to unstable grasp configurations. On the other hand, if a
sampled point was too far from the boundary, the palm of the hand might collide
with the object. The distance filter removes these unstable and collided grasp
configurations. A result of the distance filter is shown in
Fig.\ref{graspplanning}(c). (2) Near-neighbour filter. Two sampled points should
not be too near to each other.
Near sampled points increase the density of auto-planned grasp
configurations, which is unnecessary and leads to high computational cost.
We merge the sampled points in a region by representing them using a single
point. The process is done using the fixed-radius nearest neighbour algorithm.
The neighbours that fall inside the radius of a chosen sampled
point will be removed. The result of the near-neighbour filter is shown in
Fig.\ref{graspplanning}(d).

\textbf{Parallel facets:} For a parallel gripper, the contact points of the two
fingers pads must be on two parallel facets. Therefore we compute the
force-closure grasps by finding all parallel facets where the sampled points on
one facet can be projected to the inner region of the other facet along its
inverse normal direction. Some results of the parallel facets are shown in
Fig.\ref{graspplanning}(e)-(h). The contact points, their projections, and
the normals of the contact points and projections are illustrated using colored
arrows.

One sampled point together with its projection on one of its parallel facet is
called a pair of contact points. For each pair of contact points, we pose the
two finger pads on them and sample the rotation of the hand around the axis
passing through the contact pair. An example of the sampled rotations is
shown in Fig.\ref{graspplanning}(m) (ignore the color for this subsection,
the rotated hands are gripping at a pair of contact points shown in
Fig.\ref{graspplanning}(k)).

\textbf{Resistance to gravity torques:}
The resistance of a grasp to gravity torques is measured using the distance
between the contact pair and the $com$ (center of mass) of the object. Since the
object will be reoriented during manipulation, the maximum gravity torque would
be $mg|\boldsymbol{p_{com}}-\boldsymbol{p_{grp}}|$ where $m$ is the mass of the
object, $g$ is the gravity constant, $\boldsymbol{p_{com}}$ is the center of
mass, $\boldsymbol{p_{grp}}$ is the center of the contact pair. If
$|\boldsymbol{p_{com}}-\boldsymbol{p_{grp}}|$ is larger than a threshold, the contact pair is judged to be
unstable during reorient. The candidate grasp configurations at the contact pair
will be removed.

\textbf{Collision detection:}
Two levels of collision detection are used to remove collided grasps. The first
level uses the swept volumes of the finger pads during closing the gripper to
detect the collisions between finger pads and the object. The swept volumes are
modeled as two cylinders since cylinder models are invariant to rotation around
the axis passing through the contact pair and only one collision detection is
needed. Some examples of the collision detections in the first
level are shown in Fig.\ref{graspplanning}(i)-(l). The second level uses the
model of the robotic hand to remove the collisions between the whole hand and the object. The
collision detection is performed at each sampled rotation around the axis
passing through the contact pair. An example is shown in
Fig.\ref{graspplanning}(m) and (n) where the red hands indicate the collided grasp configurations. The white hands
indicate the collision-free grasp configurations. The first level of collision
detection is fast and reduces the necessity to check the collision between hands
and objects at some contact pairs in the second level. They together expedite
the collision detection process.

A fast grasp planner that plans robust grasp configurations to grasp objects of
various shapes could be implemented using the aforementioned algorithms.

\section{Using RDB to Manage the Planned Results}

The auto-planned grasps, together with the stable placements and other
pre-computed results are saved in a relational
database for reuse and analysis. RDB (Relational DataBase), rather than file
systems, is used to help process the large amount of data and their
relationships.

\begin{figure}[!htbp]
	\centering
	\includegraphics[width=.487\textwidth]{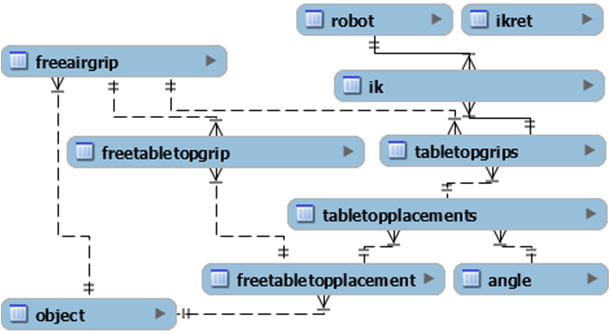}
	\caption{ERG (Entity Relation Graph) of the RDB. The database includes 10
	tables to save objects, robots, grasps, placements, and their relationships. }
	\label{rdgegg}
\end{figure}

The ERG (Entity Relation Graph) of the database is shown in Fig.\ref{rdgegg}.
The database is composed of ten tables named \bsf{object}, \bsf{robot},
\bsf{freeairgrip}, \bsf{freetabletopplacement}, \bsf{freetabletopgrip},
\bsf{angle}, \bsf{tabletopplacements}, \bsf{tabletopgrips}, \bsf{ikret}, and
\bsf{ik}, respectively.

\begin{figure*}[!htbp]
	\centering
	\includegraphics[width=\textwidth]{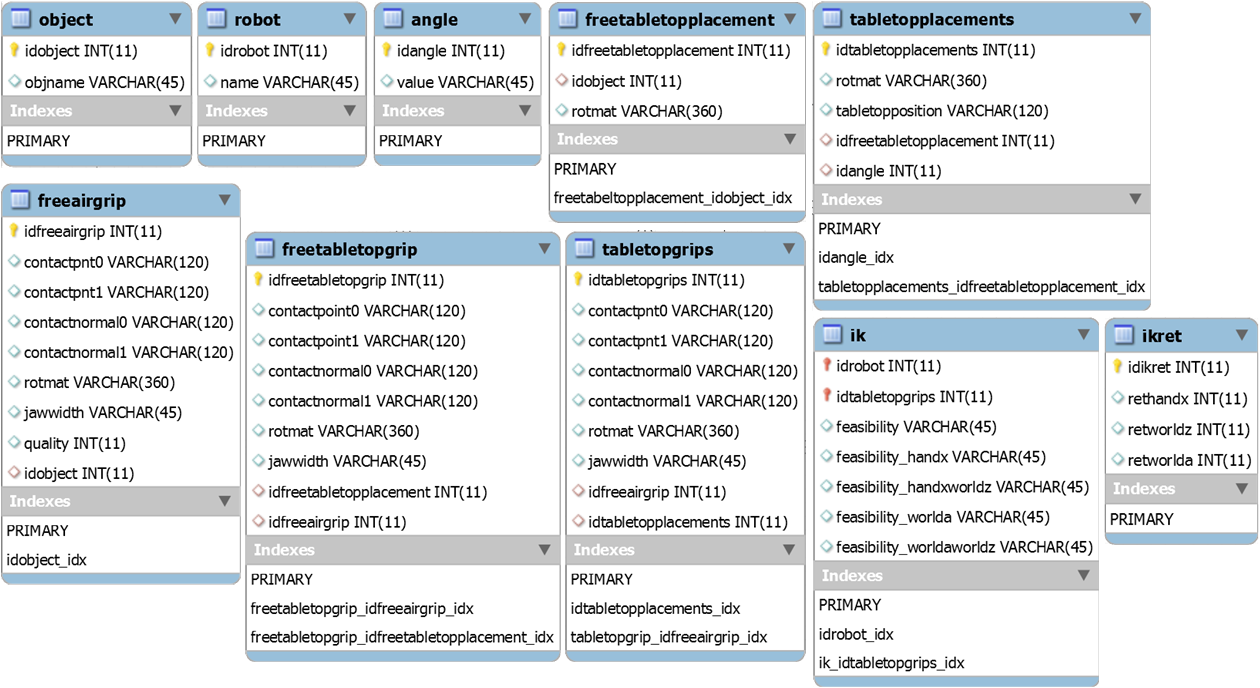}
	\caption{Primary keys, foreign keys, and columns of the tables in the RDB.}
	\label{rdgeggdetail}
\end{figure*}

The contents of the tables are shown in
Fig.\ref{rdgeggdetail}. The object and robot tables save the names of the
objects and robots. Each row of them has a primary key named
\textsf{idxxx}(\textsf{idobject} or \textsf{idrobot}) and a second column storing the name.
The \bsf{freeairgrip} table saves the grasp configurations in the local
coordinate systems of objects, without considering surrounding obstacles. The
\bsf{freeairgrip} table has a foreign key pointing to the id of \bsf{object}.
It has a 1:n relationship with \bsf{object} (Fig.\ref{rdgegg}). The
\bsf{freetabletopplacement} table saves the placements of objects on a table. It
is named free since the horizontal coordinates are always at (0,0), and the
rotation around vertical axis is always 0 (they are ready to be displaced and rotated freely). The pose of the placements are saved in
\bsf{freetabletopplacement}.\textsf{rotmat}. The
\bsf{freetabletopplacement} table has a 1:n relationship
with the \bsf{object} table and has a foreign key pointing to the id of
\bsf{object}.
The \bsf{freetabletopgrip} table saves the available grasp configurations of
\bsf{freetabletopplacement}.
Its columns include the contact points (\textsf{contactpoint0} and
\textsf{contactpoint1}), the contact normals (\textsf{contactnormal0} and
\textsf{contactnormal1}), the pose of the hand (\textsf{rotmat}), and the
opening width of the jaw (\textsf{jawwidth}). It has 1:n relationships with
\bsf{freeairgrip} and \bsf{freetabletopplacement}. The grasp configurations of
\bsf{freetabletopplacement} are based on the \bsf{freeairgrip}. We do not
re-plan them but transform the \bsf{freeairgrip} to the coordinate systems of
the \bsf{freetabletopplacement}. We re-compute the availability of the
transformed grasp configurations using collision detection, and save the
available ones to \bsf{freetabletopgrip}.
The \bsf{tabletopplacements} table also saves the placements of objects on a
table. Compared with \bsf{freetabletopplacement}, the horizontal
coordinates are not fixed to (0,0). They are discretized to specific positions.
Also, the rotation around vertical axis is discretized to specific values
saved in the \bsf{angle} table. The \bsf{tabletopplacements} table is
re-computed from the \bsf{freetabletoplacement} table, and therefore has a 1:n
relationship with
\bsf{freetabletopplacement}. It also has a 1:n relationship with the \bsf{angle}
table. The \bsf{tabletopgrips} table is similar to the \bsf{freetabletopgrip}
table. It saves the available grasp configurations of \bsf{tabletopplacements}.
The \bsf{tabletopgrips} table is re-computed using \bsf{freeairgrip} and has a
1:n relationship with \bsf{freeairgrip} and \bsf{tabletopplacements}. The
\bsf{ik} table saves the feasibility of the grasp configurations in
\bsf{tabletopgrips} with respect to specific robots. It therefore has 1:n
relationships with \bsf{tablettopgrips} and \bsf{robot}. The primary
key of \bsf{ik} is composed of two foreign keys: The id of
\bsf{robot} and the id of \bsf{tabletopgrips}. It saves the feasibility of the
grasp configurations denoted by \bsf{ik}.\textsf{idtabletopgrips}, and also
saves several other feasibility after some retraction. For example,
\bsf{ik}.\textsf{feasibility\_handx} is the feasibility of IK after retracting
hand configurations along their $x$ directions.
\bsf{ik}.\textsf{feasibility\_handxworldz} is the feasibility of IK after first
retracting hand configurations along their $x$ directions and then
along $z$ direction of the world. The retraction distances are pre-defined
and saved in the \bsf{ikret} table.

\begin{figure}[!htbp]
	\centering
	\includegraphics[width=.457\textwidth]{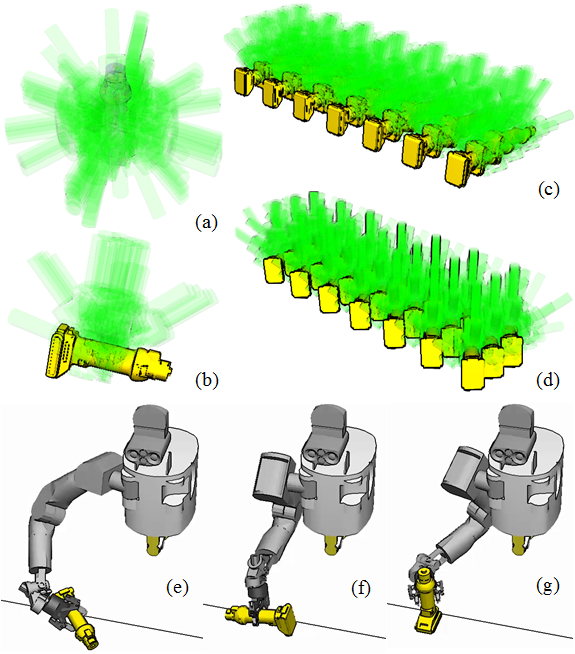}
	\caption{(a) The \bsf{freeairgrip} of an electric drill object. The
	collision-free and force-closed grasp configuration are rendered in green.
	(b) The \bsf{freetabletopgrip} of the drill at one placement. The grasp
	configurations that collide with a table supporting the placement are removed. (c)-(d) The
	\bsf{tabletopgrips} of the placement all over a table surface. (c) and (d) are
	at two rotations of the \bsf{angle} table.
	(e)-(g) Some feasible IKs with respect to a HRP5P robot.}
	\label{dbiks}
\end{figure}

Fig.\ref{dbiks} visualizes some of the tables.
Fig.\ref{dbiks}(a) and (b) show the \bsf{freeairgrip} and
\bsf{freetabletopgrip} of an electric drill object. Fig.\ref{dbiks}(c) and (d)
show two rotated \bsf{tabletopgrips}. 
Fig.\ref{dbiks}(e)-(g) show some feasible IKs with respect to a humanoid
robot with an 8-DoF (Degree of Freedom) arm (name: HRP5P).

\section{Roadmap-based Regrasp Graph}

\begin{figure*}[!htbp]
	\centering
	\includegraphics[width=.98\textwidth]{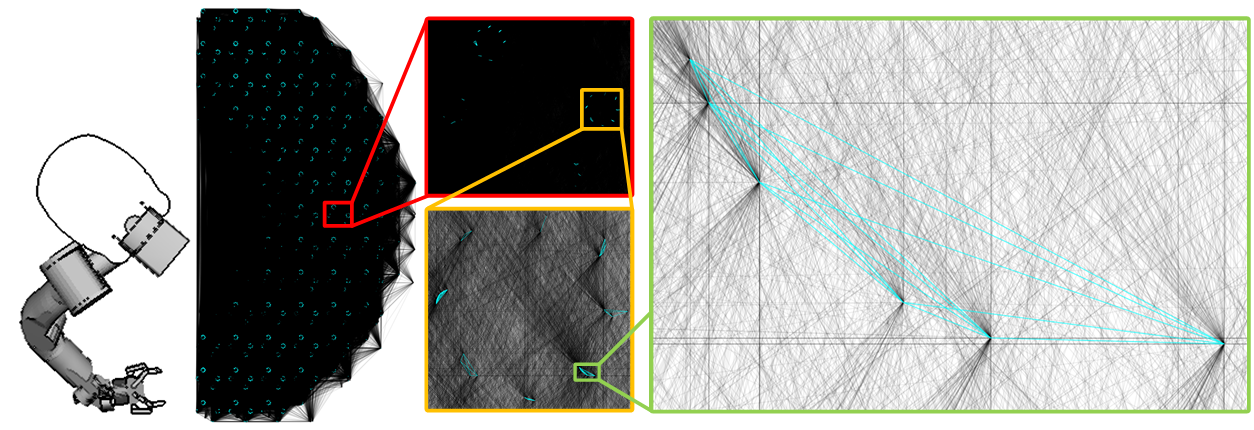}
	\caption{The regrasp graph is a roadmap built on a table in front of the
	robot. The nodes of the graph are show in the green box. Each node corresponds
	to one grasp. One component connected by the cyan edges denote the grasps of
	one discretized rotation. The eight components in the orange box indicate
	rotation is discretized into eight angles. The stable placements at a
	position on the table are represented by the cyan circles in the red box
	(four stable placements in this case).}
	\label{graph}
\end{figure*}

The data and relationships saved in the RDB make it easy to build the graph.
Using them, we can analyze the combinatorics of grasps and placements, and build
regrasp graphs on a table in front of a robot.
The graph encodes the positions all over the table like a roadmap
(Fig.\ref{graph}).

\textbf{Building the graph}
The nodes of the graph are from the \bsf{tabletopgrips} table. Each node
indicates one grasp configuration. The edges of the graph are converted from the
relationships of \bsf{tabletopplacements}, \bsf{tabletopgrips}, and \bsf{freeairgrip}.
Two nodes are connected using a transfer edge when the two rows of
\bsf{tabletopgrips} share the same \textsf{idfreeairgrip}, which means a robot
could grasp the object at one placement (one row of \bsf{tabletopplacements})
using a grasp configuration and transfer the object to another placement
(another row of \bsf{tabletopplacements}) using the same grasp configuration.
Two nodes are connected using a transit edge when the two rows of the
\bsf{tabletopgrips} share the same \textsf{idtabletopplacements}, which means a robot could release the object
grasped by a grasp configuration (one node at one end of the edge), transit to a
different grasp configuration (the node at the other end of the edge), and grasp
the object again using the second grasp configuration.

One example of the regrasp graph using \bsf{tabletopplacements}
is shown in Fig.\ref{graph}. There are 2912 placements in the graph. These
placements are at 91 positions on the table in front of a robot. At each
position, the object could be posed at 4 different stable placements with
8 discretized rotations around vertical axis. The available grasps of a
placement (a stable placement at a specific orientation) are connected to
each other using transit edges (cyan). The shared grasps of different placements
are connected to each other using transfer edges (black).

\begin{figure}[!htbp]
	\centering
	\includegraphics[width=.46\textwidth]{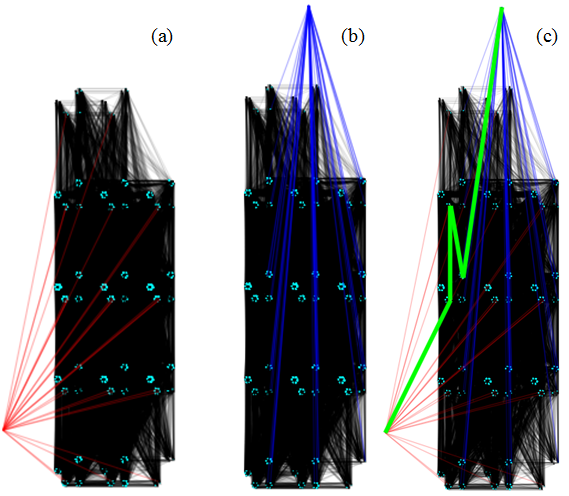}
	\caption{(a)-(b) Connecting the initial grasps (red) and goal grasps (blue) to
	the regrasp graph. (c) Searching the regrasp graph to find a sequence of
	transit and transfer grasps (green).}
	\label{searchg}
\end{figure}

\textbf{Searching the graph}
Once built, the regrasp graph can be used repeatedly for the same object. A
vision system detects the initial pose of the object. The user inputs the goal
pose of the object. The graph searching algorithm computes the available grasps
of the initial and goal poses, and connects them to the roadmap. An example is
shown in Fig.\ref{searchg}. The initial placement and their available grasps are
connected to the graph using red edges. The goal placement and their available
grasps are connected to the graph using blue edges.
The graph searching algorithm finds a path (green) from one of the initial
grasps to one of the goal grasps. There might be several
candidate paths, which could be selected using some criteria.

\section{Experiments and Experiences}

\begin{figure}[!htbp]
	\centering
	\includegraphics[width=.47\textwidth]{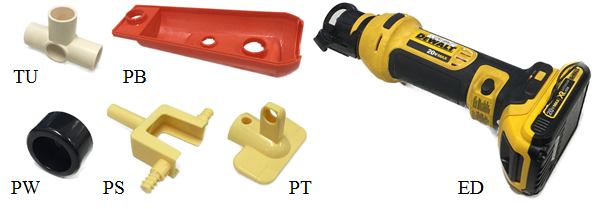}
	\caption{The objects used in the experiments.}
	\label{expobj}
\end{figure}

The proposed algorithms are validated using various objects and robots.
The objects used include (Fig.\ref{expobj}): (1) A plastic tube (TU), (2) a toy
plane body (PB), (3) a toy plane wheel (PW), (4) a toy
plane support (PS), (5) a toy plane tail (PT), and (6) an electric drill
(ED). The robots used
include: (1) HRP5P, which is a humanoid robot developed
by our institute, and (2) Kawada Nextage, which is a commercially available
industrial dual-arm robot. For each robot, a single arm is used. The processor
of our computer is Intel Xeon 2.8GHz. Its graphic card is NVIDIA Quadro M3000M.

\begin{figure}[!htbp]
	\centering
	\includegraphics[width=.47\textwidth]{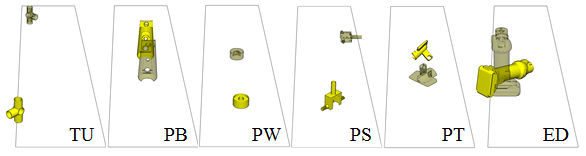}
	\caption{Some tasks. The objects in solid
	yellow are at initial poses. The objects in transparent yellow are at
	goal poses.}
	\label{tasks}
\end{figure}

The tasks used to validate the algorithms are shown in Fig.\ref{tasks}.
Robots need to reorient the objects from the poses rendered in solid
yellow to the poses in transparent yellow.

\begin{table*}[!htbp]
\centering
\renewcommand{\arraystretch}{1.2}
\caption{\label{dstc}Data size and time cost}
\begin{threeparttable}
\begin{tabular}{lllllllllll|llll}
\toprule
 \bb{obj} & \bb{\#-tri} & \bb{\#-fg} & \bb{t-fg} & \bb{\#-fp} & \bb{\#-fpg} &
 \bb{t-fp} & \bb{\#-tp} & \bb{\#-tpg} & \bb{t-tpgik$_n$}
 & \bb{t-tpgik$_h$} & \bb{gs$_n$} & \bb{gs$_h$} & \bb{nr$_n$} & \bb{nr$_h$}
 \\
 \midrule
 \midrule
 TU & 1640 & 84 & \gr{10.43$s$} & 6 & \brt{0,38,1,18,33,21} &
 \gr{6.45$s$} & 1,008 & \brt{18,648} & \bg{1,746.67$s$} & \bg{1,692.74$s$} &
 \br{5.51$s$} & \br{15.52$s$} & 1 & 3\\
 PB & 2786 & 1 & \gr{10.33$s$ }& 2 & \brt{0,1} & \gr{0.51$s$} & 336 & \brt{168}
 & \bg{4,025.69$s$} & \bg{4,069.84$s$} & \br{0.14$s$} & \br{0.17$s$} & 0 & 0\\
 PW & 936 & 312 & \gr{5.93$s$} & 2 & \brt{77,84} & \gr{12.38$s$} & 336 &
 \brt{27,048} & \bg{2,200.94$s$} & \bg{2,346.49$s$} & \br{15.82$s$} &
 \br{30.91$s$} & 0 & 1\\
 PS & 13,662 & 157 & \gr{101.76$s$} & 3 & \brt{97,44,35} & \gr{9.36$s$} & 504 &
 \brt{29,568} & \bg{16.46$s$} & \bg{16.09$s$} & \br{8.62$s$} & \br{28.01$s$} &
 2 & 2\\
 PT & 18,066 & 202 & \gr{218.72$s$} & 4 & \brt{46,64,65,47} & \gr{15.91$s$} &
 672 & \brt{37,296} & \bg{2,680.36$s$} & \bg{2,679.52$s$} & \br{11.39$s$} &
 \br{25.08$s$} & 1 & 3\\
 ED & 34,960 & 118 & \gr{700.94$s$} & 5 & \brt{106,25,21,41,44} & \gr{8.43$s$} &
 840 & \brt{39,816} & \bg{3,186.77$s$} & \bg{2,891.07$s$} &  \br{7.95$s$} &
 \br{28.91$s$} & 1 & 1\\
\midrule
\bottomrule
\end{tabular}
\begin{tablenotes}[para,flushleft]
Meanings of the abbreviations: \bb{obj} -- object name, \bb{\#-tri} -- number of
triangle meshes in the object's mesh model, \bb{\#-fg} -- number of free grasps
(without consider obstacles, see \bsf{freeairgrip} in RDB), \bb{t-fg} -- the
time needed to compute the free grasps, \bb{\#-fp} -- number of free placements
(without displacements and rotations, see \bsf{freetabletopplacement}
in RDB), \bb{\#-fpg} -- the number of available grasps for each free placement
of the object (see \bsf{freetabletopgrip} in RDB), \bb{t-fp} -- the time needed
to compute the free placements and update their available grasps, \bb{\#-tp} --
number of total placements on the table (see \bsf{tabletopplacements} in RDB),
\bb{\#-tpg} - the number of available grasps for all placements,
\bb{t-tpgik$_n$} -- the time needed to solve the IK of all grasps for Nextage,
\bb{t-tpgik$_h$} -- the time needed to solve the IK of all grasps for HRP5P,
\bb{gs$_n$} -- the time needed to connect the initial and goal grasps to the
regrasp graph and search a path for Nextage, \bb{gs$_h$} -- the time needed to
connect the initial and goal grasps to the regrasp graph and search a path for
HRP5P, \bb{nr$_n$} -- the number of regrasps needed to reorient the object by
Nextage, \bb{nr$_h$} -- the number of regrasps needed to reorient the object by HRP5P.
\end{tablenotes}
\end{threeparttable}
\end{table*}

\textbf{Computational cost}
The size of the database and the time cost of the various planning
algorithms are shown in Table.\ref{dstc}. The meanings of the abbreviations are
in the footnote of the table. The columns before the vertical separator are the
volume of the data and the general cost to prepare the data. Particularly, the
columns colored in gray are fully off-line. The columns colored in brown are flexible, depending on how
complete practitioners would like the planner to be. For a table with a fixed
height, they can be fully off-line. For tables with varying heights, they must
be recomputed. 10,000$s$ of grasps are planned and saved for regrasp planning
(see the \bb{\#-tpg} column).
The columns after the vertical separator show the specific costs of the tasks in
Fig.\ref{tasks}. The time used to compute the IK-feasible grasps at initial and
goal poses, connect the grasps to the regrasp graph, and search the graph for
the two robots are shown in the \bb{gs$_n$} (Nextage) and \bb{gs$_h$} (HRP5P)
columns. They are in interactive time using the shown data volume. 
The values of \bb{gs} depend on \bb{\#-fpg} and \bb{\#-tpg} (the columns colored
in light purple). For example, the PW task has more \bb{\#-fpg} and \bb{\#-tpg}
and therefore costs more. The PB task has few \bb{\#-fpg} and \bb{\#-tpg} and
costs less than 0.2$s$ (the success rate is lower). The times of regrasps are
shown in the \bb{nr$_r$} (Nextage) and \bb{nr$_h$} (HRP5P) columns. The Nextage
robot has better kinematic design to reorient the objects in some tasks: It used
1, 0, and 1 regrasps to reorient objects TU, PW, and PT. In contrast, the HRP5P
robot used 3, 1, and 3 times of regrasps to reorient them.
Also, one arm of HRP5P has 8 DoFs and costs more to do
connecting and searching (including
the shoulder and waist; One arm of Nextage has 7 DoFs).

\begin{figure*}[!htbp]
	\centering
	\includegraphics[width=\textwidth]{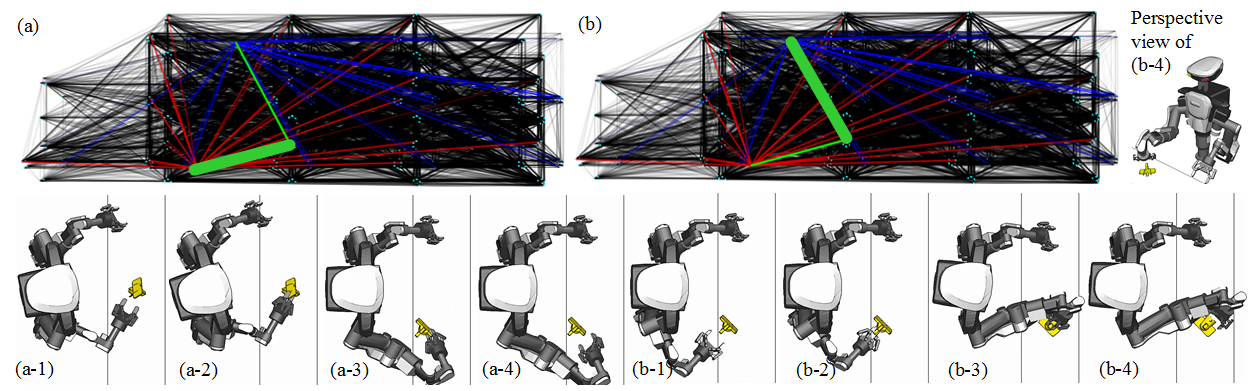}
	\caption{A sequence of robot poses and grasp configurations planned for the
	Nextage robot to perform the task shown in Fig.\ref{tasks}(PT). (a)-(b) The
	regrasp graph and the found path. (a-1)-(a-4) correspond to the first
	segment of the path shown in (a). (b-1)-(b-4) correspond to the second
	segment shown in (b). A perspective view of (b-4) is shown in the upper-right
	corner. The kinematic structure of the robot enables it to finish the task
	using one regrasp. It did fewer regrasps than HRP5P as the pose in (b-4) is not
	reachable by HRP5P (compared with Fig.\ref{rgptailhrp5}).}
	\label{rgptailnxt}
\end{figure*}

\begin{figure*}[!htbp]
	\centering
	\includegraphics[width=\textwidth]{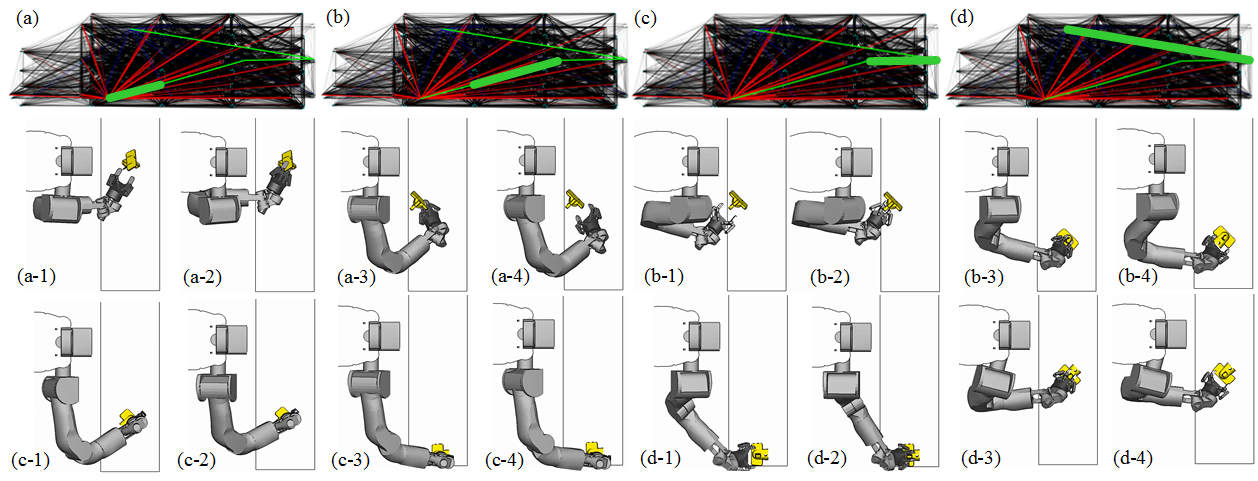}
	\caption{A sequence of robot poses and grasp configurations planned for the
	HRP5P robot to perform the task shown in Fig.\ref{tasks}(PT). The task is the
	same as Fig.\ref{rgptailnxt}. The HRP5P robots used three regrasps to finish
	the task. Likewise, the segment in (a) is detailed in (a-1)-(a-4). The segment
	in (b) is detailed in (b-1)-(b-4), etc.}
	\label{rgptailhrp5}
\end{figure*}

\textbf{Planned sequences}
Some planned manipulation sequence for the two robots to reorient object PT
(task Fig.\ref{tasks}(PT)) are shown in Fig.\ref{rgptailnxt} and
Fig.\ref{rgptailhrp5}. The Nextage Robot has more
flexibility than the HRP5P robot in this workspace. It used one regrasp to
finish the task. In contrast, the HRP5P robot used three regrasps. The
perspective view of Fig.\ref{rgptailnxt}(b-4) is shown in the upper left corner of
Fig.\ref{rgptailnxt}. The robot is at a pose which is not reachable by HRP5P.

\textbf{Experiences}
During the implementation, we carefully designed several parameters including:
(1) Density of surface sampling, (2) stability of a placement, (3) resistance
to torque caused by gravity, etc. The density of surface sampling is crucial to
the number of automatically planned grasps and computational feasibility. The
stability is essential to reduce accumulated errors and achieve high success
rate during regrasp. The resistance to torque and gravity is important to
certainty of grasps and stability during reorient. To make them general, we
computed the density of sampling using the size of mesh surfaces, computed the
stability of a placement using the ratio between the height of $com$ and the
distance to the boundary of the supporting polygon, and filtered the resistance
to torque caused by gravity by adding thresholds to the distance between the
contact center and the $com$ of the object. These strategies are adaptive to
varying model geometry (see Fig.\ref{tasks} and
the \bb{\#-tri} column of Table.\ref{dstc}), but they cannot be adaptive to
varying physical properties like Coulumb coefficients, uneven density, etc.
Dealing with uncertainties caused by these physical properties is an open problem.

\section{Conclusions}

In this paper, we developed intelligent algorithms for robots to reorient
objects using 10,000$s$ grasps. We developed robust
grasp planning algorithms to plan the grasps and used RDB to manage the
automatically planned data. These data were reused during regrasp planning to
build regrasp roadmaps and find robot-pose and grasp-configuration sequences to
reorient objects. Experiments showed the developed algorithm with the support of
the database can reuse 10,000$s$ of grasps to reorient objects at various poses
in interactive time. We conclude that (1) the grasp planning algorithms are
robust to find more grasps, (2) the relational database
successfully manages the large amount of data generated by planning
algorithms, and (3) the algorithms leverage modern computational ability to
challenge the relationships and the combinatorics of the data. They are
applicable to various robots and objects.

\section*{Acknowledgment}
The paper is based on results obtained from a project commissioned by the
New Energy and Industrial Technology Development Organization (NEDO).

\ifCLASSOPTIONcaptionsoff
  \newpage
\fi
\bibliographystyle{IEEEtran}
\bibliography{reference}

\end{document}